\def\Eqref#1{Equation~\ref{#1}}
\def\1{\bm{1}}

\def\vx{{\bm{x}}}

\def\mA{{\bm{A}}}

\def\mH{{\bm{H}}}
\def\mI{{\bm{I}}}
\def\mJ{{\bm{J}}}

\def\mW{{\bm{W}}}

\DeclareMathAlphabet{\mathsfit}{\encodingdefault}{\sfdefault}{m}{sl}
\SetMathAlphabet{\mathsfit}{bold}{\encodingdefault}{\sfdefault}{bx}{n}

\def\emW{{W}}

\documentclass{article}

\usepackage[english]{babel}
\usepackage{amssymb,amsfonts,bm,amsmath,algorithm, algpseudocode}
\usepackage{caption}
\usepackage{subcaption}
\usepackage{booktabs}
\usepackage{color}
\usepackage{xcolor}
\usepackage{natbib}
\usepackage[letterpaper,top=2cm,bottom=2cm,left=3cm,right=3cm,marginparwidth=1.75cm]{geometry}

\usepackage{amsmath}
\usepackage{graphicx}
\usepackage{mathtools}
\usepackage[colorlinks=true, allcolors=blue]{hyperref}
\title{Regularizing Deep Neural Networks with Stochastic Estimators of Hessian Trace}
\author{Yucong Liu \\ University of Chicago \and Shixing Yu \\ University of Texas Austin \and Tong Lin\footnote{Corresponding author} \\ Peking University}

\begin{document}
\maketitle

\begin{abstract}
In this paper, we develop a novel regularization method for deep neural networks by penalizing the trace of Hessian. 
This regularizer is motivated by a recent guarantee bound of the generalization error. We explain its benefits in finding flat minima and avoiding Lyapunov stability in dynamical systems. We adopt the Hutchinson method as a classical unbiased estimator for the trace of a matrix and further accelerate its calculation using a dropout scheme. Experiments demonstrate that our method outperforms existing regularizers and data augmentation methods, such as Jacobian, Confidence Penalty, Label Smoothing, Cutout, and Mixup.
\end{abstract}

\section{Introduction}
\label{}
Deep neural networks (DNNs) are developing rapidly in recent years. The effectiveness of DNNs has been widely demonstrated in multiple areas including % image classification~\cite{}, object detection~\cite{}, machine translation~\cite{}, text generation~\cite{}, speech recognition~\cite{}.
% and are widely used in many fields such as 
reflection removal \citep{removal}, dust pollution \citep{dust}, building defects detection \citep{building}, cities and urban development \citep{city}. 
In literature, researchers have been exploring on different  architectures design of neural networks, e.g. residual connections~\citep{resnet}, batch normalization~\citep{ioffe2015batch} or special design for different type of activation functions~\citep{hendrycks2016gaussian, xu2015empirical} to boost the performance.
However, among various learning problems, over-fitting on training data has been consistently demonstrated to be a essential problem that greatly affected the generalization ability of deep learning models. To alleviate the overfitting problem and to achieve better generalization, regularization has long been developed to penalize on the complexity of the system.

There are two classical regression methods reducing model complexity in linear models, namely Ridge Regression \citep{ridge} and Lasso Regression \citep{lasso} that has profound influence on following regularization schemes.
Inspecting from its core methodology, they are also called $L_{2}$ and $L_{1}$ regularization. $L_{2}$ regularization has the effect of shrinkage while $L_{1}$ regularization can be beneficial to both shrinkage and sparsity, with a special focus on the sparse structure. 
From a Bayesian perspective, they can be interpreted as posterior estimation given different prior, where ridge regression uses normal distribution and Lasso regression follows a Laplacian prior respectively.

As classical Lasso and Ridge Regression for linear models can be elegantly and effectively applied to the regularization of neural networks, the situation here is more complex than in linear models. So sophisticated regularization methods have been developed. 
An intuitive idea is starting from penalizing the magnitude of weights.
The most widely used method is Weight-Decay \citep{weightdecay}, which is a technique to shrink parameters to 0 before updating by gradient.
\citet{l2wd} also showed that $L_{2}$ regularization and Weight-Decay are not identical. 
Dropout \citep{dropout} is another method to avoid over-fitting by reducing co-adapting between units in neural networks. Dropout has inspired a large body of work studying its effects (\cite{wager2013dropout,helmbold2015inductive, wei2020implicit}). After dropout, various regularization schemes can be applied additionally, such as Label Smoothing \citep{label} and Confidence Penalty \citep{confidence}. For more literature review, see section \ref{relatedwork}. 

Most existing methods only consider about the scale of parameters and the shape of output confidence. None of them takes the local properties near the minima into consideration, such as Lyapunov stability or sharpness. Our work focus on such properties of better minima and how to find that one by gradient descent. This is the essential difference between our regularizer and existing ones.

%\shixing{Explain about why previous methods failed.(Never consider about the sharpness of the minima.)}

Our contributions are summarized as follows:
\begin{itemize}
    \item We develop a novel regularization scheme by penalizing the Hessian trace. This regularization method is directly inspired by a recent generalization bound about the Jacobian norm and the Hessian trace. The Hessian trace can also be connected to flat minima with good generalization performance and stability analysis of a nonlinear dynamical system. 
    \item We develop a stochastic Hessian trace estimation algorithm for efficient implementation in practice.
    \item Experimental results demonstrate that our new regularization method outperforms other existing approaches on both vision and language tasks.
\end{itemize}

\section{Related Work}
\label{relatedwork}

There are many regularization methods in previous work. Label Smoothing \citep{label} estimates the marginalized effect of label-dropout and reduces over-fitting by preventing a network from assigning a full probability to each training example. Similarly, Confidence Penalty \citep{confidence} prevents peaked confidence distributions, leading to better generalization. A network appears to be overconfident when it places all probability on a single class in the training set, which is often a symptom of over-fitting. DropBlock \citep{dropblock} is a structured form of dropout: instead of dropping out independent random units, contiguous regions are dropped from a feature map of each layer. 

Data augmentation methods are also used in practice to improve the model’s accuracy and robustness when training neural networks. Cutout \citep{cutout} is a data augmentation method where parts of the input examples are zeroed out, in order to
focus more on less prominent features for generalizing to masked regions. Mixup \citep{mixup} extends the training distribution by incorporating the prior knowledge that linear interpolations of feature vectors should lead to linear interpolations of the associated targets.

\citet{jacobian} first proposed Jacobian regularization, a method focusing on the norm of Jacobian matrix with respect to input data. It was proved that generalization error can be bounded by the norm of Jacobian matrix. Besides that, Jacobian matrix shows improved stability of the model predictions against input perturbations according to Taylor expansion. \citet{hoffman2019robust} showed that Jacobian regularization enlarges the size of decision cells and is practically effective in improving the generalization error and robustness of the models. To simplify calculation, stochastic algorithm of Jacobian regularization was also proposed.

Motivated by Jacobian regularization, we consider the generalization error and stability of the model concerning the Hessian matrix. We propose Hessian regularization with corresponding stochastic algorithms. We compare our Hessian regularization with other methods and demonstrate promising performance in experiments. The main idea to estimate the trace of the Hessian matrix is Hutchinson Method \citep{trace} which was also discussed by \citet{pyhessian}. We make an improvement by designing a new probability distribution to dropout parameters, which decreases time consumption obviously without losing generalization. 

Hessian information is powerful tool used on analyzing the property of neural networks. \citet{adahessian} designed AdaHessian, a second order stochastic optimization algorithm. A Hessian-Aware Pruning method \citep{yushixing} was developed to find insensitive parameters in a neural network, and a Neural Implant technique was also proposed to alleviate accuracy degration.  However, their methods are static in essence, whereas we focus on dynamical motion of parameters in parameter space. \citet{yiyangdenage} also proposed a Hessian regularization. They focused on the layerwise loss landscape via the eigenspectrum of the Hessian at each layer. We start from different perspectives of generalization error bound and dynamical system of parameters. Our experiments also shows better results than \citet{yiyangdenage}'s method.

The literature of \citet{1997flat}, \citet{large}, \citet{sharp} observed that the flatness of minima of the loss landscape results in good generalization. In contrast, sharp minima lead to poorer generalization. Concretely, sharp minima can be characterized by a significant number of large positive eigenvalues in the Hessian matrix which tend to generalize poorly. They found that small-batch gradient methods converge to flat minimizers, which can be characterized by having numerous small eigenvalues of the Hessian matrix.  Our newly designed regularization method enables the neural network to find flat minima for achieving better performance.

\section{Motivations}
In this section, we start with notations and definitions in subsection \ref{Sec:3.1}. Then we introduce a generalization error bound in subsection \ref{Sec:3.2}, which motivates us to constrain the Hessian trace. After that, we explain the benefit on searching flat minima by penalizing Hessian trace in subsection \ref{Sec:3.3}. Furthermore, the connection between the Hessian trace and Nonlinear Stability Analysis is shown in subsection \ref{Sec:3.4}. Finally we define the Hessian regularization in subsection \ref{Sec:3.5}.

\subsection{Notations and Definitions}\label{Sec:3.1}
Suppose there is a hypothesis function $h: \mathcal{X} \rightarrow \mathcal{Y}$ output an target $y \in \mathcal{Y}$, given an input feature vector $x \in \mathcal{X}$. Denote the joint distribution of $x$ and $y$ as $P(x,y)$. Sample set $\mathcal{S}$ consists of $n$ instances ${\ (x_{1},y_{1}),\ldots ,(x_{n},y_{n})}$ drawn i.i.d. from 
$P(x,y)$. With this sample set, we want to use a DNN model $f(x;\omega)$ to approximate $h(x)$, where $x$ is input data and $\omega$ is trainable parameters. Let $\ell$ be a non-negative real-valued loss function, where $\ell(f(x;\omega),y)$ measures the difference between the prediction $f(x;\omega)$ and the ground-truth $y$.

The empirical loss of $f(x;\omega)$ associated with the sample set is the average of loss for each sample, which is defined as
\begin{equation}
    \ell_{emp}(f)=\hat{\mathbb{E}}[\ell(f(x;\omega), y)]=\frac{1}{n}\sum_{(x_{i}, y_{i})\in \mathcal{S}} \ell(f(x_{i};\omega),y_{i}),
\end{equation}
and the expected loss of $f(x)$ is the expectation of loss under the joint distribution $P$, defined as
\begin{equation}
    \ell_{exp}(f)=\mathbb{E}[\ell(f(x;\omega),y)]=\mathbb{E}_{(x,y) \sim P}[\ell(f(x;\omega),y)].
\end{equation}
Then the difference between $\ell_{emp}(f)$ and $\ell_{exp}(f)$
is called generalization error:
\begin{equation}
     GE(f)=|| \ell_{exp}(f)-\ell_{emp}(f) ||.
\end{equation}

\subsection{Generalization Error Bound}\label{Sec:3.2}
In this subsection, we introduce a recent generalization error bound involving Hessian trace. 

\citet{wei2020implicit} showed a generalization error bound of linear models with cross-entropy loss of $M$ classes. Let $\mW$ is the weight matrix, $\mu(\mW) \coloneqq \hat{E} [\left\|\mJ\right\|_{2}]$ and $v(\mW) \coloneqq \hat{E} [tr(\mH)]$, where $\mJ$ denotes the Jacobian matrix and $\mH$ denotes the Hessian matrix. Thus, $\mu(\mW)$ is the average Jacobian matrix norm and $v(\mW)$ is the average Hessian trace over samples. Then, with probability $1-\delta$ over the training examples, for all weight matrices $\mW$ satisfying the norm bound $|| \mW^{T} ||_{2,1} \leq A$, the following bound holds:
\begin{equation}
    \mathbb{E}[\bar{\ell}]-1.01\hat{\mathbb{E}}[\bar{\ell}]
    \lesssim
    \frac{(A\mu(\mW))^{\frac{2}{3}}(\theta B)^{\frac{1}{3}}}{n^{\frac{1}{3}}}
    +\frac{A\sqrt{Bv(\mW)\theta}}{\sqrt{n}}
    +\frac{BA^{2}\theta}{n(\log^{2}(\frac{BA^{2}\theta}{v(\mW)n})+1)}
    +\zeta.
\end{equation}
Here with some fixed bound $B > 0$, we define
\begin{equation}
    \begin{aligned}
        &\bar{\ell}=min\{\ell,B\},
        \\
        &\left\|\mW\right\|_{2,1}
        =\sum_{j}\sqrt{\sum_{i}(\emW_{ij}^{2})},
        \\
        &\theta=\ln^{3}(nM)\mathop{\max}_{i}\left\|x_{i}\right\|_{2}^{2},
        \\
        &\zeta=\frac{B(\ln(1/\delta)+\ln\ln n)}{n}.
    \end{aligned}
\end{equation}

So one can guarantee good generalization when both the norm of the Jacobian matrix and the trace of the Hessian matrix are small. On one hand, when learning with gradient descent, we want to find a local or global minima of the loss function. Naturally, at minima the gradient should be zero and the norm of the Jacobian matrix is small near minima. So gradient descent helps us to ensure the norm of the Jacobian matrix is small.

On the other hand, gradient descent only considers first-order derivative, it can not constrain the trace of the Hessian matrix, which is composed of the second-order derivative. From this aspect, it is necessary to add a restriction on the Hessian trace when updating parameters by gradient descent. Though the above generalization error bound holds for linear models, it is natural to generalize from linear models to DNN models: each layer of a DNN model can be viewed as a linear model except for the non-linear activation functions.

Thus, the generalization error bound indicates that restriction on Hessian trace is essential for training a DNN model. 

\subsection{Flat Minima}\label{Sec:3.3}
Previous work \citet{1997flat,large,sharp} show that flat minima have good generalization. A flat minima is a large connected region in parameter space where the error remains approximately constant. The Bayesian argument suggests that flat minima correspond to “simple” networks and prevent over-fitting. 

We start by writing Taylor expansion of loss function $\ell$ at a minima $\omega_{*}$. 
Since at convergence of local minima $\omega_{*}$, the first-order derivative of $\ell$ is 0, then we have 
$$
\ell({\omega}) = \ell({\omega_{*}}) + (\omega - \omega_{*})^{T} \mH (\omega - \omega_{*}) + o(\|\omega - \omega_{*}\|^{2}),
$$
where $\mH$ denotes the Hessian matrix of $\ell$ with respect $\omega_{*}$.

Each eigenvalue of $\mH$ indicates the extent of how $\ell$ changes with minor perturbation on $\omega$ on the corresponding eigenvector direction. 
Then smaller eigenvalue suggests that the minima is flat in the corresponding eigenvector direction. 
If the Hessian $\mH$ at $\omega_{*}$ has all eigenvalues with smaller magnitudes, then it is a relatively flatter minima. Hence, penalizing on eigenvalues of Hessian can help lead to a flat minima. 

Motivated by the theoretical analysis mentioned above, it's necessary to develop a regularizer using second-order information, especially the eigenvalues of Hessian. However, Hessian is a symmetric matrix with a tremendous amount of elements, i.e. $O(n^2)$ where $n$ is the parameter size commonly occurs to be million level in NNs. Calculating the full hessian of size $O(n^2)$ is intractable neither in a computation perspective nor in the memory constraint of modern computation resources. Even though we have the Hessian, the computation of eigenvalues is also complicated. Thus, we need to find a scalar index that distills the information of Hessian, which is cheap to compute and has similar properties as eigenvalues. As the sum of eigenvalues, the Hessian trace is a good option. Penalizing Hessian trace implies that the magnitudes of Hessian eigenvalues can be constrained, enhancing the chance of finding flat minima.

\subsection{Linear Stability Analysis}\label{Sec:3.4}
In this subsection, we focus on the optimization process of parameter $\omega$ during the training process and the local properties around the minima.  

The optimization process through stochastic gradient descent can be regarded as a motion process in the parameter space, from the landscape of initialization to convergence at a local or global minima. 
At each discrete step, the parameter as a high-dimensional vector takes the gradient as a moving direction. Then gradient descent can be formulated as a series of discrete updates:
\begin{equation}
    \omega_{t+1}=\omega_{t}-\eta g_{t},
\end{equation}
where $\omega_{t}$ is the parameter position at step $t$, $\eta$ is learning rate and gradient $g_{t}$ is written as:
$$
g_t = g_t(\omega_t,x) = \frac{d\ell(f(x;\omega_t),y)}{d\omega_t}.
$$

Considering learning rate as the discrete time interval of updating weights, i.e. $\Delta t = \eta$, and denoting $\Delta w$ as the parameter update, we have:
\begin{equation}
    \frac{\Delta \omega}{\Delta t}=-g(\omega,x).
\end{equation}
Under the assumption that time interval is small enough, approximately we can reformulate the discrete update rule into a contiunous form:
\begin{equation}\label{ode}
    \frac{d\omega}{dt}=-g(\omega,x).
\end{equation}
Thereafter, with an initial condition, we have the complete trajectory of parameter point based on Ordinary Differential Equation (ODE) theory. The process of gradient descent is transformed to a Nonlinear Dynamical System (NDS), which allows us to leverage the basic theory of linear stability analysis for developing new methods.

According to nonlinear dynamical systems, the updating gradient $-g$ in Equation~\ref{ode} is referred to as rate function. Denoted as the equilibrium points $\omega_{*}$ of an NDS, the minima are such parameters where the rate function vanishes $-g(\omega_{*})=0$. 
If any solution starting near an equilibrium point leaves the neighborhood of $\omega_{*}$ as $t \to \infty$, then $\omega_{*}$ is called \textbf{asymptotically unstable}, while if all solutions starting within the neighborhood approach $\omega_{*}$ as $t \to \infty$ then the equilibrium is called \textbf{asymptotically stable}. \citet{lyapunovstablity} gave more rigorous definition and discussion, known as \textbf{Lyapunov Stability Theory}.
Intuitively, during neural network training, it's beneficial to have the optimizer find unstable minima along the moving trajectory in the parameter space. In this way, it's easier for the weight vector to jump out of the local minima and to search for a better descent direction.

In an NDS, the stability of an equilibrium point is highly correlated with the Jacobian matrix denoted as $\mJ$, which is the Jacobian matrix of rate function $-g$ w.r.t. to the weights $\omega$. 
The Theorem of Lyapunov
Stability~\citep{mathmodel} states that at the equilibrium point $\omega_{*}$, if all eigenvalues of the Jacobian $\mJ(\omega_{*})$ have real parts less than zero, then $\omega_{*}$ is Lyapunov stable. 

Back to gradient descent, the Jacobian matrix in a NDS is exactly the negative of the Hessian matrix in a neural network. 
Denoting the Hessian matrix of the loss function $\ell$ with respect to parameter $\omega$ as $\mH$, we have 
$$
\mH = \frac{\partial}{\partial \omega}\left( \frac{\partial \ell}{\partial \omega} \right) = \frac{ \partial g}{\partial \omega} = -\mJ(\omega_{*}).
$$
The Hessian $\mH$ is a real symmetric matrix and all its eigenvalues are real numbers. Since the Hessian trace is the sum of the eigenvalues, penalizing the Hessian trace helps decrease the eigenvalues to some extent. Then it increases the eigenvalues of the Jacobian matrix in the NDS, avoiding the eigenvalues to be negative. Thus, the constraint on the Hessian trace has a inclination  to weaken the Lyapunov stability of the minima.

Despite reducing stability sounds bad, it is beneficial for escaping the local minima towards finding global minima. 
A Lyapunov stability point is referred to as an easily-converged equilibrium point. Once the parameter $\omega$ gets close to the Lyapunov stable equilibrium point, $\omega$ has little probability of escaping from this equilibrium point.
Thus, falling to a stable equilibrium point will trap the optimization from finding a better local minima.
Similar to the idea of the Confidence Penalty, regularizing on the Hessian trace can help the model to rethink and to escape these `easily converged' equilibrium points. Avoiding these Lyapunov stable points can make the gradient method find a better equilibrium point for generalization.

\subsection{Hessian regularization}\label{Sec:3.5}
We define the Hessian regularization term as 
\begin{equation}
    tr(\mH_{\ell,\omega}).
\end{equation}
It's the trace of second derivative of empirical loss $\ell$ with respect to parameters $\omega$.
Then, we define a new loss with our Hessian regularization as
\begin{equation}
    Loss = \ell_{emp}(f) + \lambda \cdot tr(\mH_{\ell,\omega}),
\end{equation}
where $\lambda$ controls the strength of the Hessian regularization. Based on  previous analysis on the generalization error bound, flat minima and Lyapunov stability, the Hessian regularization is expected to improve generalization performance.

\section{Algorithms}
As discussed before, there are more than millions of parameters in a typical DNN. So the calculation of the Hessian matrix is difficult. Thus, we introduce two efficient stochastic algorithms to estimate Hessian trace, SEHT-H in subsection \ref{Sec:4.1} and SEHT-D in subsection \ref{Sec:4.2}.

\subsection{Hutchinson Method}\label{Sec:4.1}
 Hutchinson Method \citep{trace} is an unbiased estimator for the trace of a matrix. Let $\mA$ be an $n \times n$ symmetric matrix with $tr(\mA) \neq 0$. Let $\sigma$ be a random vector whose entries are i.i.d Rademacher random variables ($Pr(\sigma_{i} = \pm 1) = \frac{1}{2}$), then $\sigma^{T} \mA \sigma$ is an unbiased
estimator of $tr(\mA)$, based on the following equation:
\begin{equation}\label{trace}
    tr(\mA)=tr(\mA\mI)
    =tr(\mA\mathbb{E}[\sigma\sigma^{T}])
    =\mathbb{E}[tr(\mA\sigma\sigma^{T})]
    =\mathbb{E}[tr(\sigma^{T}\mA\sigma)]
    =\mathbb{E}[\sigma^{T}\mA\sigma].
\end{equation}

In this paper, we consider the trace of Hessian matrix $\mH$, which is the second derivative matrix. Since the Rademacher random vector is irrelevant to network parameters, 
$$
\frac{d \sigma}{d \omega} = 0.
$$ 
Then we expand the expression of Hutchinson estimator as follow:
\begin{equation}\label{eq:hut}
    \sigma^{T}\mH\sigma
    =\sigma^{T}\frac{d}{d\omega}\left(\frac{dl}{d\omega}\right)\sigma
    =\sigma^{T}\left[\frac{d}{d\omega}\left(\frac{dl}{d\omega}\right)\cdot \sigma+\frac{dl}{d\omega}\cdot\frac{d\sigma}{d\omega}\right]
    =\sigma^{T}\frac{d}{d\omega}\left(\frac{dl}{d\omega}\cdot \sigma \right).
\end{equation}

Based on \Eqref{eq:hut}, we can estimate the Hessian trace by calculating the gradient of loss $g_{\omega}=\frac{dl}{d\omega}$ and the gradient of $\frac{dl}{d\omega}\cdot \sigma$. We do not need the prohibitive computation of the whole Hessian matrix. The fast second-order information estimation only includes two inner products and two gradients. We refer to the Hutchinson stochastic estimator of Hessian trace as SEHT-H, presented in Algorithm~\ref{alg:SEHT-H}. In practice, we only focus on the weight parameters in each layer of DNN and ignore the bias parameters.

\begin{algorithm}[tb]
   \caption{SEHT-H}
   \label{alg:SEHT-H}
\begin{algorithmic}
   \State {\bfseries Input:} $n$-dimensional gradient $g$
   \State {\bfseries Output:} Estimation of $tr(\mH)$
   %\REPEAT
   % \State Initialize $noChange = true$.
   \For{$i=1$ {\bfseries to} $maxIter$}
   \State $\sigma \sim Rademacher(n)$ 
   %\tcc*{n-dim vector with each element sampled from Rademacher distribution.}
    
    \State $v = g \cdot \sigma$ %\tcc*{inner product}
    
    \State $h = \frac{dv}{d\omega}$ %\tcc*{derivative of $\vv$}
    
    \State $t = \sigma^{T} \cdot h$ %\tcc*{inner product}
    
    \State $trace \mathrel{+}= t$ \;
   %\IF{$x_i > x_{i+1}$}
   %\State Swap $x_i$ and $x_{i+1}$
   %\State $noChange = false$
   %\ENDIF
   \EndFor
   \State {\bfseries Return:} $ \frac{trace}{maxIter}$
   %\UNTIL{$noChange$ is $true$}
\end{algorithmic}
\end{algorithm}

Even though SEHT-H is a stochastic algorithm which reduces considerable amount of computational overhead, it is still not efficient enough due to the great number of parameters of a neural network. Hence, we propose to modify the pipeline of SEHT-H based on the basic idea of Dropout. The Dropout method boosts its computation speed and make it more efficient for neural network training.

\subsection{Dropout Method}\label{Sec:4.2}
Inspired by Dropout ~\citep{dropout}, we then propose a stochastic parametric method to accelerate SEHT-H.
In Dropout, every node in a neural network has a probability $p$ to be ignored in the training process to reduce co-adaptations. Thus, in each training iteration, only a random sub-network of the original network is used.
Intuitively, in the process of Hessian trace calculation, not all the parameters are necessary to be considered.
it would be much faster if we only use a small subset of the network parameters during each calculation.
% In our Hessian regularization, we want to lower the trace $tr(\mH)=\sum_{i} \frac{\partial^{2}\ell}{\partial \omega_{i} \partial \omega_{i}}$, the sum of diagonal elements of Hessian matrix. Based on the idea of Dropout, 
Thus, we ignore some parameters when constraining Hessian trace $tr(\mH)$. %, since reducing the partial sum of diagonal elements can have a large chance to reduce the total sum. 
The sum of the selected subset of diagonal elements is denoted as $\tilde{tr}(\mH)$. Considering the layer structures of neural networks, the process of randomized parameter selection can be divided into two steps: 
$(i)$ randomly select layers in neural network with probability $p_{1}$, and
$(ii)$ randomly select parameters in the selected layers with probability $p_{2}$. 
In other words, when carrying out Hessian regularization, we ignore layers with probability $1-p_{1}$, ignore parameters in the selected layers with probability $1-p_{2}$. In our experiment, we simply set $p_{1}=p_{2}$.

Following the basic idea of Hutchinson algorithm, we want to obtain Hessian trace without the heavy calculation of Hessian matrix.
To extend the algorithm from full-parameter domain to partial-parameter domain, here we define a new probability distribution $Q(p)$. If $\vx \sim Q(p)$, then 
$$
\begin{aligned}
    &Pr(\vx = \pm 1) = p, \\
    &Pr(\vx = 0) = 1-2p.
\end{aligned}
$$ Then, supposing that $\sigma$ is a random vector whose entries are i.i.d. random variables following the $Q$ distribution, we have
\begin{equation}
    \mathbb{E}[\sigma\sigma^{T}|\; \text{fix  the positions of} \; 0 \; \text{in} \; \sigma] = \tilde{\mI}.
\end{equation}
Here $\tilde{I} = \displaystyle \text{diag}(0,1)$ is a diagonal matrix with diagonal elements equal to 0 or 1. Notice that, the non-diagonal entries of $\mathbb{E}[\sigma\sigma^{T}]$ are all zero, because for $i \neq j$,
$$ 
\mathbb{E}[\sigma_{i}\sigma_{j}] = 1 \times 2p^{2} + (-1) \times 2p^{2} = 0.
$$

Then similar to \Eqref{trace}, if we fix the positions of 0 in $\sigma$, we have unbiased estimator of the partial sum of diagonal elements:
\begin{equation}
    \tilde{tr}(\mA)=tr(\mA\tilde{\mI})
    =tr(\mA\mathbb{E}[\sigma\sigma^{T}])
    =\mathbb{E}[tr(\mA\sigma\sigma^{T})]
    =\mathbb{E}[\sigma^{T}\mA\sigma].
\end{equation}
We can expand the expression same as \Eqref{eq:hut} and transform the calculation process into two inner products and two gradients. This efficient method with random selected subset of parameters for calculating Hessian trace is presented in Algorithm~\ref{alg:SEHT-D}. We name it as SEHT-D.

\begin{algorithm}[tb]
   \caption{SEHT-D}
   \label{alg:SEHT-D}
\begin{algorithmic}
   \State {\bfseries Input:} probability $p$, parameter $\omega$ in selected layers,  and corresponding $n$-dim gradient $g$
   \State {\bfseries Output:} Estimation of $\tilde{tr}(\mH)$
   \For{$i=1$ {\bfseries to} $maxIter$}
   \State $\sigma \sim Q(p)$ 
   %\tcc*{n-dim vector with each element sampled from Rademacher distribution.}
    
    \State $v = g \cdot \sigma$ %\tcc*{inner product}
    
    \State $h = \frac{dv}{d\omega}$ %\tcc*{derivative of $\vv$}
    
    \State $t = \sigma^{T} \cdot h$ %\tcc*{inner product}
    
    \State $trace \mathrel{+}= t$ \;
   \EndFor
   \State {\bfseries Return:} $ \frac{trace}{maxIter}$
\end{algorithmic}
\end{algorithm}

SEHT-D makes Hessian regularization tractable for neural network training.
Thus, in practice, we mainly provide results of SEHT-D. 
Compared with other regularization methods, including Label Smoothing and Confidence Penalty, our Hessian regularization shows improved test performance.

\section{Experiments}

We evaluate the proposed Hessian regularization method (abbreviated as SEHT) on the tasks of image classification and language modeling respectively, which are commonly used to measure the effectiveness of regularization.

For comparison across a variety of datasets, we adopt several existing regularization methods and data augmentation methods, e.g., Confidence Penalty~\citep{confidence}, Label Smoothing~\citep{label}, Cutout~\citep{cutout}, MixUp~\citep{mixup}. 
To implement existing methods on our backbone model for fair comparison, we have done a thorough hyperparameter search to extract the best performance of each regularization method.

\subsection{Image Classification}
\subsubsection{CIFAR-10}

The CIFAR-10 dataset consists of 60000 instances of 32x32 colour images in 10 classes, with 6000 images per class. 
There are 50000 training images and 10000 test images.
On CIFAR-10 experiment, we use ResNet-18 \citep{resnet} as the backbone neural network. 

For all models, we set coefficient of weight decay of $5\times 10^{-4}$. 
We set learning rate 0.01, batch size 32, momentum 0.9 and all models are trained 200 epochs with Cosine Annealing~\citep{loshchilov2016sgdr}.
For Jacobian regularization, we set number of projections $n_{proj} = 1$ and weight parameter $\lambda_{JR} = 0.01$. 
For DropBlock, $block\_size = 7$ and $keep\_prob = 0.9$. 
We perform a grid search over weight values \{0.001, 0.005, 0.01, 0.05, 0.1\}
and select 0.01 for Label Smoothing, 0.001 for Confidence Penalty. Cutout size is set to $16\times 16$ based on the validation results mentioned by \citet{cutout}. For Mixup, we set $\alpha = 1.0$ according to \cite{mixup}'s setting, which results in interpolations of $\lambda$ uniformly distributed between 0 and 1.

For the proposed Hessian regularization SEHT-D, we perform a grid search over weight values \{0.0001, 0.001, 0.01, 0.1\} and finally select 0.001, testing with probability value 0.01 and 0.05.
We provide results on different possibilites on the the max number of iterations that updates the trace calculation, defined in Algorithm \ref{alg:SEHT-H}. It is denoted as maxIters and with a set candidate numbers in \{1, 5, 10\}. 
We also test SEHT-H (maxIter=5), with weight value in \{0.0001, 0.001, 0.01, 0.1\} and select 0.001 for SEHT-H. 

The top-1 accuracy is reported in Table \ref{resnet18} with standard errors over 5 runs.

\begin{table}[t]
\caption{Results of ResNet-18 on CIFAR-10 over 5 runs}
\label{resnet18}
\vskip 0.15in
\begin{center}
\begin{small}
\begin{sc}
\begin{tabular}{lcccr}
\toprule
Model & Test Accuracy$(\%)$ \\
\midrule
Baseline with Weight-Decay  &   $94.00\pm0.24$  \\
Jacobian \citep{hoffman2019robust}    &   $89.23\pm0.52$  \\
DropBlock \citep{dropblock}	&   $89.23\pm0.22$  \\
Confidence Penalty \citep{confidence}	&   $94.40\pm0.02$  \\
Label Smoothing \citep{label} &   $94.40\pm0.07$  \\
cutout \citep{cutout} &   $94.02\pm0.22$  \\
mixup \citep{mixup}	&   $95.39\pm0.13$  \\
Vanilla \citep{bregman} &   $93.6$  \\
Middle Network Method \citep{yiyangdenage}	&   $88.13 \pm0.12$    \\
MM + FRL \citep{fine}    &   $95.33\pm0.12$  \\
Lookahead \citep{look}  &   $95.23\pm0.19$  \\
\midrule
SEHT-D(maxIter=1, prob=0.01)    &   $95.37 \pm0.09$    \\
SEHT-D(maxIter=10, prob=0.01)   &   $95.42 \pm0.11$    \\
SEHT-D(maxIter=10, prob=0.05)	&   $95.49\pm 0.06$   \\
SEHT-H(maxIter=5)   &   $\bf95.59\pm0.06$  \\

\bottomrule
\end{tabular}
\end{sc}
\end{small}
\end{center}
\vskip -0.1in
\end{table}

\begin{figure}
\begin{subfigure}{0.5\textwidth}
\includegraphics[width=1.1\linewidth]{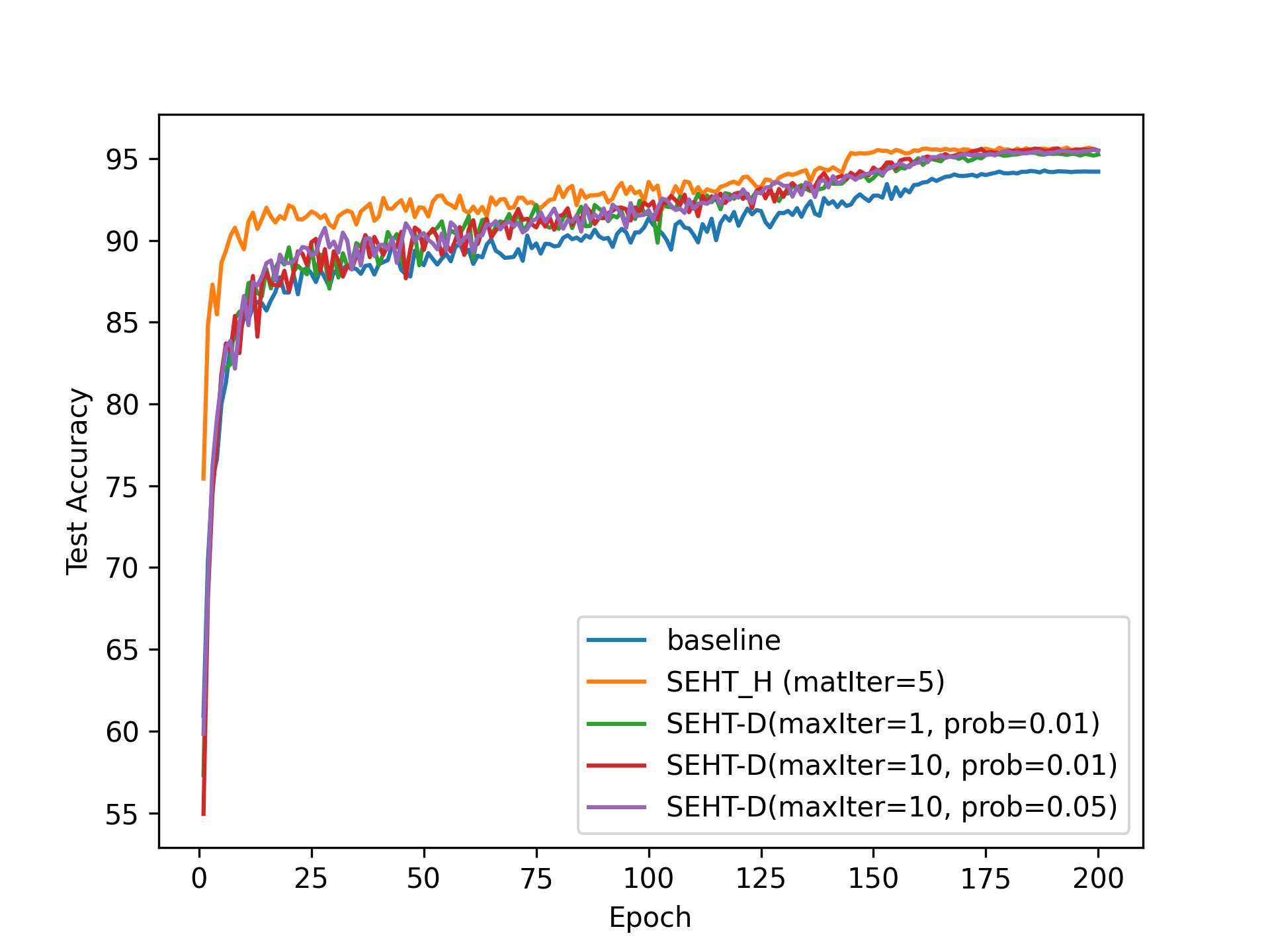}
\caption{Convergence on CIFAR-10}
\end{subfigure}
\hspace*{\fill}% this is optional just to fix the horizontal space between the two images
\begin{subfigure}{0.5\textwidth}
\includegraphics[width=1.1\linewidth]{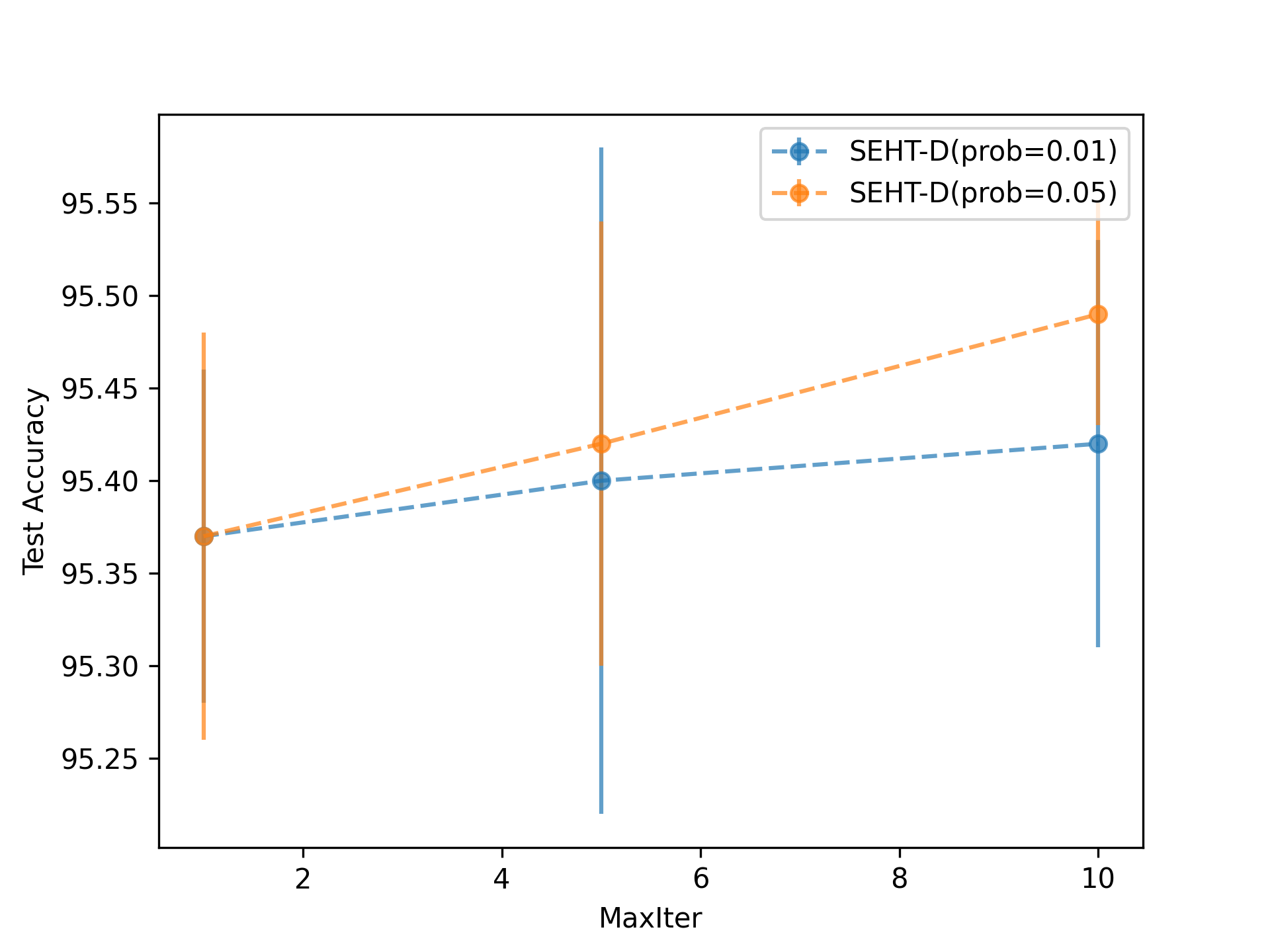}
\caption{Test Accuracy of SEHT-D on CIFAR-10}
\end{subfigure}
\caption{Performance of SEHT-D on CIFAR-10} \label{fig1}
\end{figure}

Firstly, SEHT demonstrates its superb effectiveness in improving generalization by providing the highest test accuracy on CIFAR-10. Furthermore, both SEHT-H and SEHT-D have relatively small standard errors, which means they are stable in their performance.

Our first observation falls on comparison with classical regularization and data augmentation methods. it is surprising to find out that Jacobian regularization and Dropblock method have even worse performance than the baseline with Weight-Decay. Confidence Penalty and Label Smoothing are two popular regularization methods that penalize the prediction distribution widely adopted for training a model with better generalization ability. They indeed surpass the baseline performance but with the limited improvement of 0.4\%. The best results among this category is data augmentation with MixUp~\citep{mixup} that provides a 95.39\% accuracy, but our method still consistently provide competitive or better results under different hyperparameter settings.

Recent work such as Vanilla~\citep{bregman}, MM + FRL~\citep{fine}, Lookahead~\citep{look} provide inferior accuracy to the proposed SEHT-D. In \citet{yiyangdenage}'s experiment, they got test accuracy 88.13\% on CIFAR-10 with ResNet-18, which is much worse than our result: 95.37\% with SEHT-D (maxIter=1, prob=0.01) and 95.49\% with SEHT-D (maxIter=10, prob=0.05). Moreover, their improvement is only 0.02\% for full-network and 0.10\% for middle-network. Our Hessian regularization method improves the model 1.37\% on test accuracy with SEHT-D (maxIter=1, prob=0.01) and improves 1.49\% on test accuracy with SEHT-D (maxIter=10, prob=0.05), which are much more than their improvement.
These results demonstrate the effectiveness of penalizing the Hessian trace.

Secondly, we provide different setting combinations of the maxIter parameter and the prob parameter. A trade-off can be found between computational cost and performance. SEHT-H achieves the best performance regardless of training efficiency. On the other hand, SEHT-D(maxIter=1, prob=0.01) only requires $1.2\times$ training time of the baseline and SEHT-D (maxIter=1, prob=0.05) costs only $1.3\times$ of the baseline, while SEHT-H is much slower. The larger maxIter and prob are, the more time it would take in the algorithm. As a result, SEHT-D achieves a balance between performance and time efficiency in this experiment.

% Our first observation is on the computational cost of the training with SEHT-D (maxIter=1, prob=0.01). SEHT-D(maxIter=1, prob=0.01) only requires $1.2\times$ training time of the baseline. 

Finally, the convergence results are is provided in Figure \ref{fig1}, where we compared the convergence speed and test accuracy on SEHT-D and SEHT-H with different parameters. 
It's interesting to notice that although setting different maxIter and prob can lead to different convergence trajectories, they all fall into a similar range of accuracy that clearly diverges from the baseline. This also provides evidence that SEHT may find a better minima by jumping out of stable equilibrium points.

In short, our SEHT-D and SEHT-H converge faster and better than the baseline. When increasing maxIter and prob, our SEHT-D shows better test accuracy but costs more time.

\subsubsection{CIFAR-100}
CIFAR-100 dataset is similar to the CIFAR-10 dataset, except that the target space are separated into 100 classes. 

As widely adopted on CIFAR-100, Wide Residual Networks (WRNs) are also used in our experiments as the backbone neural network. Specifically, WRN-28-10 is used with depth 28 and fixed widening factor of 10. 
For all models,  Weight Decay is set to $5\times 10^{-4}$, batch size to 32, momentum to 0.9, and models are trained 200 epochs. The learning rate is initially set to 0.1 and is scheduled to decrease by a factor of 5 at 60, 120, and 160 epochs. 
Different from CIFAR100, we set the Dropout probability to be 0.3 suggested by \citet{WRN}'s cross-validation since it's more difficult to learn 100 classes than CIFAR10. Cutout size of 8 × 8 pixels is used according to \citet{cutout}'s validation results. For mixup, we keep with $\alpha = 1.0$. A grid search over weight values \{0.0001, 0.001, 0.01, 0.1\} is applied for Label Smoothing, Confidence Penalty and SEHT. Then a weight value of 0.1 is used for all these three methods. We report the averaged accuracy with standard error over 5 random initialization and the results are presented in Table~\ref{tab:cifar100}.
% We also combine these three methods with Dropout. 0.0001 works best for confidence penalty with Dropout and 0.001 without Dropout. 0.001 works best for label smoothing with Dropout and 0.0001 without Dropout. 0.01 works best for SEHT with Dropout and 0.001 without Dropout. We report the mean and standard error of the mean over 5 random initialization.

In this experiment, our SEHT-D((maxIter=1, prob=0.01) method shows better results on both top-1 accuracy and top-5 accuracy, improving 4.92 and 2.52 respectively. SEHT-D((maxIter=1, prob=0.05) improves 5.70 and 2.48 respectively, out preform all other methods tested.  When testing together with Dropout, our SEHT-D has lower accuracy, which means it may have not good combination with Dropout.

On CIFAR-100, the proposed SEHT-D method shows the best results on both top-1 accuracy and top-5 accuracy compared to other methods. Separately, SEHT-D with prob=0.01 achieves the best top-5 accuracy of 95.00\% with a 0.61\% improvement than Label Smoothing
while SEHT-D with prob=0.05 outperforms Label Smoothing on top-1 accuracy drastically with around 1\% improvement. The latter one has  smaller standard errors on both top-1 accuracy and top-5 accuracy, indicating that SEHT-D (maxIter=1, prob=0.05) yields consistent good performance.

\begin{table}[t]
\caption{Results of WRN-28-10 on CIFAR-100}
\label{tab:cifar100}
\vskip 0.15in
\begin{center}
\begin{small}
\begin{sc}
\begin{tabular}{lcccr}
\toprule
Model   &   Top-1 Acc   &   Top-5 Acc \\
\midrule
Baseline  & $74.61 \pm 0.52$  & $92.48 \pm 0.40$  \\
Confidence Penalty  & $77.15 \pm 1.54$  & $93.97 \pm 0.66$    \\
Label Smoothing     & $79.38 \pm 0.26$  & $94.39 \pm 0.33$    \\
cutout  & $76.70 \pm 0.79$  & $93.72 \pm 0.40$  \\
mixup   & $78.38 \pm 0.31$  & $94.37 \pm 0.31$  \\
\hline
SEHT-D(maxIter=1, prob=0.01)   & \underline {$ 79.53 \pm 0.72$}  & $\bf 95.00 \pm 0.24$  \\
 SEHT-D(maxIter=1, prob=0.05)   &  $\bf 80.31 \pm 0.33$  & \underline{$ 94.96 \pm 0.05$}  \\

% \hline
% Dropout &   $73.13 \pm 1.00$    &   $91.89 \pm 0.52$    \\
% Confidence Penalty + Dropout    &   $77.69 \pm 0.16$    & $94.32 \pm 0.46$   \\ 
% Label Smoothing + Dropout   & $78.55 \pm 1.10$  & $94.23 \pm 0.34$  \\
% SEHT-D(maxIter=1, prob=0.01) + Dropout   & $77.75 \pm 0.37$ & $94.38 \pm 0.09$ \\
\bottomrule
\end{tabular}
\end{sc}
\end{small}
\end{center}
\vskip -0.1in
\end{table}

\subsection{Language Modeling}
\subsubsection{Wiki-text2}

The Wiki-Text language modeling dataset is a collection of over 100 million tokens extracted from the set of verified Good and Featured articles on Wikipedia.

A 2-layer LSTM \citep{LSTM} is applied as the backbone model. The size of word embeddings is 512 and the number of hidden units per layer is 512. The LSTM model is trained for 40 epochs, with batch size 20, gradient clipping 0.25, and Dropout ratio 0.5. The Dropout ratio is searched from \{0, 0.1, 0.2, 0.3, 0.4, 0.5\} and  0.5 is chosen for best performance. The initial learning rate is tuned from \{0.001, 0.01, 0.1, 0.5, 1, 10, 20, 40\} and decreases by a factor of 4 when the validation error saturates and selects 20 to be the best. Parameters are initialized from a uniform distribution $\left[ {-0.1, 0.1} \right]$. We perform the same grid search over weight values \{0.001, 0.005, 0.01, 0.05, 0.1\} for Label Smoothing, Confidence Penalty, and SEHT. The weight value of 0.01 works best for all these three methods. The probability value of 0.05 is found to outperform the value of 0.01 for Hessian regularization. We report the mean and standard error of perplexity over 5 random initialization in table~\ref{tab:lstmwiki}.

In this experiment with LSTM backbone, SEHT-D obtains the best test perplexity and Label Smoothing shows the best validation perplexity. SEHT-D improves the model by 0.79 on test perplexity. Confidence Penalty performs only slightly better than the baseline method.

\begin{table}[t]
\caption{Results of LSTM on Wiki-Text2 over 5 runs (lower is better)}
\label{tab:lstmwiki}
\vskip 0.15in
\begin{center}
\begin{small}
\begin{sc}
\begin{tabular}{lcccr}
\toprule
Model   &   Valid ppl   &   Test ppl \\
\midrule
Baseline   & $101.82 \pm 0.16$    & $95.65 \pm 0.10$ \\ 
Confidence Penalty  & $101.39 \pm 0.16 $ & $95.57 \pm 0.06 $  \\
Label Smoothing & $\bf 99.58 \pm 0.06$ & $95.03 \pm 0.30 $    \\
SEHT-D(maxIter=1, prob=0.05) & $ 100.69 \pm 0.27$  & $\bf 94.86 \pm 0.26$\\
\bottomrule
\end{tabular}
\end{sc}
\end{small}
\end{center}
\vskip -0.1in
\end{table}

For the 2-layer GRU \citep{GRU} model as the backbone, the same hyperparameter search is performed. The size of word embeddings is 512 and the number of hidden units per layer is 512. We run every algorithm for 40 epochs, with batch size  20, gradient clipping 0.25, Dropout ratio 0.3, and initial learning rate 20. Parameters are initialized from a uniform distribution $\left[ {-0.1, 0.1} \right]$. The weight value is set to be 0.05 for Label Smoothing, 0.005 for Confidence Penalty, and 0.001 for our Hessian regularization, after the grid search over \{0.001, 0.005, 0.01, 0.05, 0.1\}. Finally, the probability value is 0.01 in Hessian regularization. We repeat each method over 5 random initialization and results are presented in Table ~\ref{tab:wikitext}.

The Hessian regularization method has both the best validation perplexity and the best test perplexity, improving by 2.83 and 2.61 respectively compared with the baseline method. Confidence Penalty surpasses Label Smoothing on GRU model. Label Smoothing also shows better results than baseline.

\begin{table}[t]
\caption{Results of GRU on Wiki-Text2 over 5 runs (lower is better)}
\label{tab:wikitext}
\vskip 0.15in
\begin{center}
\begin{small}
\begin{sc}
\begin{tabular}{lcccr}
\toprule
Model   &   Valid ppl   &   Test ppl \\
\midrule
Baseline    & $119.04 \pm 2.38$    & $111.64 \pm 1.87$ \\ 
Confidence Penalty  & $116.40 \pm 0.08 $ & $109.27 \pm 0.03 $  \\
Label Smoothing & $117.47 \pm 0.24$ & $ 110.46 \pm 0.45 $    \\
SEHT-D(maxIter=1, prob=0.01) & $\bf 116.21 \pm 0.31$ & $\bf 109.03 \pm 0.15$\\
\bottomrule
\end{tabular}
\end{sc}
\end{small}
\end{center}
\vskip -0.1in
\end{table}

Our experiments on Language Modelling demonstrate that all these three regularization methods can improve models, while our SEHT-D achieves the best performance.

\section{Conclusion and Future Work}
We propose a new regularization method named Stochastic Estimators of Hessian Trace (SEHT). Our method is motivated by a guarantee bound that a lower trace of the Hessian can result in a better generalization error. It can help escape Lyapunov stable points and find flat minima. To simplify computation, our method implements two versions, SEHT-H and SEHT-D. Our experiments show that SEHT-D and SEHT-H yield promising test performance. Particularly, the SEHT method achieves $95.59\%$ accuracy on CIFAR-10 with resnet-18, outperforming classical regularization methods like Label Smoothing and Confidence Penalty.

Future works can focus on the following aspects: i)
a faster estimator on Hessian trace; ii) convergence analysis by gradient descent with second order constraint; iii) generalization error bound for multi-layer neural networks.

\bibliographystyle{abbrvnat}
\bibliography{main.bib}

\end{document}